# Hierarchical automatic lane-changing motion planning: from self-optimum to local-optimum

Yang Li, Linbo Li*, Daiheng Ni, Wenxuan Wang

Yang Li
Ph.D., Candidate
Key Laboratory of Road and Traffic Engineering of Ministry of Education,
Tongji University, China,
4800 Cao'an Road, Shanghai, 201804,
E-mail: cc960719@tongji.edu.cn

Linbo Li (corresponding author)
Ph.D., Associate Professor
Key Laboratory of Road and Traffic Engineering of Ministry of Education,
Tongji University, China,
4800 Cao'an Road, Shanghai, 201804
E-mail: llinbo@tongji.edu.cn

Daiheng Ni
Ph.D., Professor
Civil and Environmental Engineering,
University of Massachusetts Amherst, Massachusetts 01003, USA
E-mail: ni@engin.umass.edu

Wenxuan Wang
Ph.D., Candidate
Key Laboratory of Road and Traffic Engineering of Ministry of Education,
Tongji University, China,
4800 Cao'an Road, Shanghai, 201804,
E-mail: wenxuanwang@tongji.edu.cn

**Abstract**
In order to minimize the impact of lane change (LC) maneuver on surrounding traffic environment, a hierarchical automatic LC algorithm that could realize local optimum has been proposed. This algorithm consists of a tactical layer planner and an operational layer controller. The former generates a local-optimum trajectory. The comfort, efficiency, and safety of the LC vehicle and its surrounding vehicles are simultaneously satisfied in the optimization objective function. The later is designed based on vehicle kinematics model and the Model Predictive Control (MPC), which aims to minimize the tracking error and control increment. Combining macro-level and micro-level analysis, we verify the effectiveness of the proposed algorithm. Our results demonstrate that our proposed algorithm could greatly reduce the impact of LC maneuver on traffic flow. This is reflected in the decrease of total loss for nearby vehicles (such as discomfort and speed reduction), and the increase of traffic speed and throughput within the LC area. In addition, in order to guide the practical application of our algorithm, we employ the HighD dataset to validate the algorithm. This research could also be regarded as a preliminary foundational work to develop locally-optimal automatic LC algorithm. We anticipate that this research could provide valuable insights into autonomous driving technology.

# 1 INTRODUCTION

Numerous research results indicate that the advent of AVs (Autonomous Vehicles) could significantly enhance traffic safety, improve traffic efficiency, alleviate traffic congestion, and reduce fuel consumption [1, 2]. As the first city, California has formulated regulations for road testing of AVs, has attracted world's top companies to conduct research, development and road testing of AVs. The 2020 Autonomous Driving Road Test Data released by DMV [3] showed that Waymo and Crusie have conducted nearly 630,000 and 770,000 miles of testing. The corresponding MPI (Miles Per Intervention, the average number of miles traveled between every two manual interventions) is about 0.033 and 0.035 respectively. It is foreseeable that high-level AV will soon appear in daily life, and will co-exist with HV (Human-driving Vehicles) in the near future or even further.

One of the indispensable components of AV technology is the lateral maneuver research, which is a very challenging undertaking that requires exploration of solution spaces to achieve competing goals of safety, mobility, and environmental factors[4]. Generally speaking, the research on lateral maneuver research can be roughly divided into: modeling the decision-making process of LC [1, 4], the impact of LC on surroundings[5], the duration of LC [6-8], and the execution process of LC maneuver [5, 9-16]. Since this paper concentrates on the execution process, we will mainly focus on reviewing the literature which are closely related to our research theme.

The execution of LC maneuver involves the research of LTP (LC trajectory planning) and LTT (LC trajectory tracking) algorithm. When AV has made the LC decision and is about to execute LC maneuver, under the LTP and LTT algorithm, the AV would turn the steering wheel and gradually drive towards the target lane. More specifically, LTP algorithm calculates a designed LC trajectory in advance. LTT algorithm controls the vehicle to drive along this trajectory until it arrives the center-line in the target lane as shown in Figure 1.This research direction mainly revolves around these issues: (1) what is the mathematical equation form of the LC trajectory curve? (2) what factors should we consider in order to obtain the corresponding parameters? (3) how to control the vehicle to accurately follow the planned trajectory? Over the past decades, a considerable amount of efforts has been made. Existing research on LTP could be mainly divided into analytical method [12-14], artificial potential field method [15, 16], and data-driven method [10, 11]. The analytical method sets the trajectory equation in advance, and takes the needs of the LC vehicle as the optimization objective, and solves the optimal lane-changing trajectory [10-17]. Data-driven method usually refers to the method of using the machine learning or deep learning algorithm, which aims to extract LC dynamics from massive data instead of describing the nature of things[10, 11]. Recently, many scholars have tried to introduce the artificial potential field into the LTP algorithm. Artificial potential field method regards the various elements of driving environment, such as road edges, static obstacles, and moving obstacles as a potential energy field [10, 12, 14]. The vehicle tries to find a trajectory with the lowest total potential field. This paper will focus on reviewing and employing the first method, since it has high reliability and flexibility without being limited by various situations.

Up to now, the most commonly-used mathematical equations are quintic polynomial equation[13, 15, 17-19], cubic polynomial equation[11], sine(cosine) curve equation, trapezoidal curve equation, Bezier curve equation[10], etc. The problem of determining the values of corresponding parameters are generally transformed into an optimization problem, which often takes driving comfort, efficiency and safety into account. After obtaining the final form of the planned curve, the vehicle would track this curve under the control of the controller. This may involve the determination of the vehicle model (for example, the kinematic model, the kinetics model, or a more sophisticated model), and the controller (for example, the PID method, MPC method, etc.). Moreover, many scholars consider the influence of the uncertainty of the external system on the controller, and conduct the robustness analysis of the control system [16, 20].

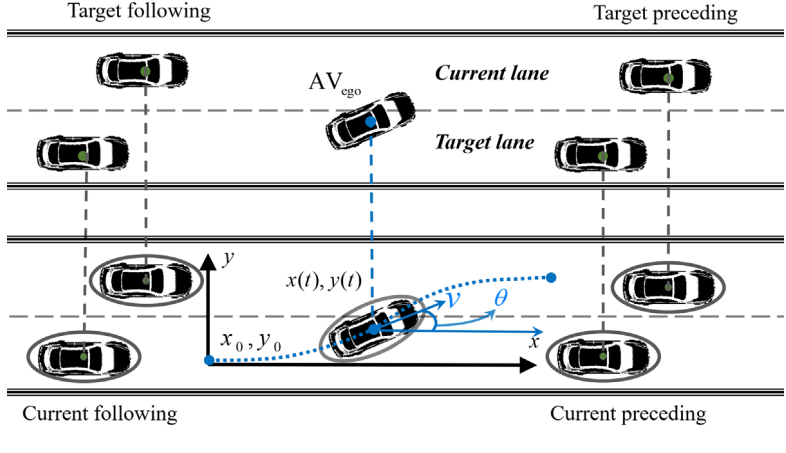

**Figure 1 The schematic diagram of LC of the autonomous vehicle (the red dot line represents the LC trajectory of the autonomous vehicle)**

Based on the V2V communication, a dynamic automated LC algorithm was proposed in [12], which consists of LTP and LTT algorithm. Driver's safety, comfort and efficiency are simultaneously satisfied in the time-based quintic polynomial function. In order to analyze the possible collision points, a rectangular collision boundary of the subject vehicle has been established in [21]. Another dynamic LTP algorithm is proposed in [20], which is composed of the trajectory decision, trajectory generation, and starting-point determination module, to respond to the changes of the state of surrounding vehicles. Personalized driving style is incorporated in the quintic polynomial function so as to meet driver's personalized LC needs in [22]. A multi-vehicle cooperative automated LTP algorithm is proposed in [23]. The cooperative safety spacing model is proposed to guarantee and improve the safety of the multiple involvement of vehicles at the same time. Due to the involvement of multiple vehicle at the same time, cooperative safety spacing model is proposed to improve the safety, and the prediction of leading vehicle is integrated in [14]. The collision area is defined to tolerate the disturbances and uncertainties of the surroundings. The generated trajectory is not allowed to cross the collision area, and thus obtaining the final optimal trajectory. A hybrid LC trajectory planning method is established, which has the strength of the sampling and optimization methods in [19].

This paper aims to find a LC trajectory that could achieve local-optimum rather than self-optimum. It is widely known that LC maneuver often leads to generation of traffic shock wave in the target lane, which has negative impacts on both travel time and traffic safety and causes congestion. Although existing research has achieved fruitful results, they often ignore the impact of LC maneuver on surrounding traffic, especially the vehicles behind in the target lane. Only the loss of the LC vehicle is considered in the optimization objective function, while overlooking the loss caused by the traffic shock wave to surrounding vehicles. Consequently, the planned trajectory might indeed be optimal for the LC vehicle. However, this may not be optimal for other surrounding vehicles, and the overall loss in the LC area might also not minimal. Therefore, in this paper, we establish a locally optimal LTP algorithm from the perspective of all vehicles within the LC area. Under this algorithm, not only the safe completion of LC maneuver is ensured, but also the impact of the LC maneuver on surrounding vehicles is minimized. We demonstrate through comprehensive numerical simulation how this algorithm may improve the overall performance of traffic flow within the LC area.

The significance of this paper is not only limited to the establishment of a novel algorithm, but more importantly is the application of the concept of local optimization for the LC maneuver. In real traffic environment, individual drivers' behaviors are often random, and they often take actions to maximize their own interests at all times. These actions may often be optimal for individuals, but not for traffic flow in their vicinity. These self-optimum actions often lead to low throughput, poor driver comfort, or even traffic accidents. Traffic management agencies typically alleviate traffic congestion passively by means of Ramp Metering [24], Variable Speed Control[25], etc. Although these measures could address traffic shock waves and congestion that are recurrent in nature, they are unable to tackle issues that are spontaneous such as those caused by lane change maneuvers. On the other hand, with the advent of self-driving vehicles, AV could autonomously perceive and obtain the status of their surrounding vehicles. Through adjusting its real-time control strategy, it becomes possible for AV to actively dissipate or eliminate traffic shock waves from the source as much as possible. This may transform traffic control from conventional passive and global means to active and individualized control. This research can also be regarded as a preliminary foundational work for applying this concept to develop locally-optimum automatic LC algorithm.

## 2 PROBLEM STATEMENT

To better illustrate our research, we present the schematic diagram of LC scenario as shown in Figure 2. Typically, the LC maneuver could be divided into two stages. One is the from the decision to the execution point (stage 1). After the driver decides to change lanes, he needs to search for a suitable gap acceptance on the target lane. This stage is mainly reflected in the driver's own internal activities. At the same time, there are also situations where the driver directly executes the LC maneuver after making the decision ($t_0 \leq t_{start}$). The other stage is the execution process of LC (stage 2). The driver begins to turn the steering wheel, and gradually arrives at the target lane, which could be viewed as external. At time $t_{start}$, there are a certain number of HVs behind the AV on the target lane. Due to the LC of AV, all vehicles behind the AV on the target lane will slow down so as to keep a safe CF (Car-following) distance. This inevitably form a shock wave that propagates backwards, reducing the efficiency and comfort of all vehicle behind in the target lane. These HVs all have made a certain degree of compromise. As elaborated above, existing mandatory LC algorithms only consider the safety, comfort and efficiency of the subject vehicle when solving the "optimal" equation curve. The driving experience of the rear vehicles on the target lane is often overlooked. This is reflected in the poor comfort and low efficiency of these vehicles. Therefore, the LC trajectory obtained in the existing research might be optimal for the AV, but not for the vehicles within the LC area. How to smooth out the traffic shock wave caused by the LC maneuver is of great interest. This may help to improve the operational condition of the traffic flow within the LC area. Considering that it is impossible for this paper to cover all LC scenarios, we will focus on establishing the algorithm under mandatory LC scenario, thus shedding light on more complex driving environment.

To facilitate the subsequent discussion, it is necessary to explain the scope and the assumption of this paper. (1) for special reasons, the AV has to change lanes. This may be due to the lower speed or the occurrence of traffic accident of the preceding vehicle on the current lane. This leads to a poor driving experience of AV, which causes the AV to perform LC. (2) the SAE Inter-national has defined six distinct levels of AV. As the level increases, the automation system has more control over the vehicle. We assume the AV is with SAE Level 4/5 automation that could drive autonomously[26]. The AV could sense real-time status of surrounding vehicles, such as speed, acceleration, and location. (3) the research of the process from the decision point to the execution point is not covered in this paper (stage 1). Admittedly, the determination of the execution point is also an indispensable part of automated LC algorithm. Nevertheless, this paper will concentrate on the motion planning part (stage 2). The determination of the local-optimal execution point, and the combination of these two stages together will be carried out in the follow-up of this research.

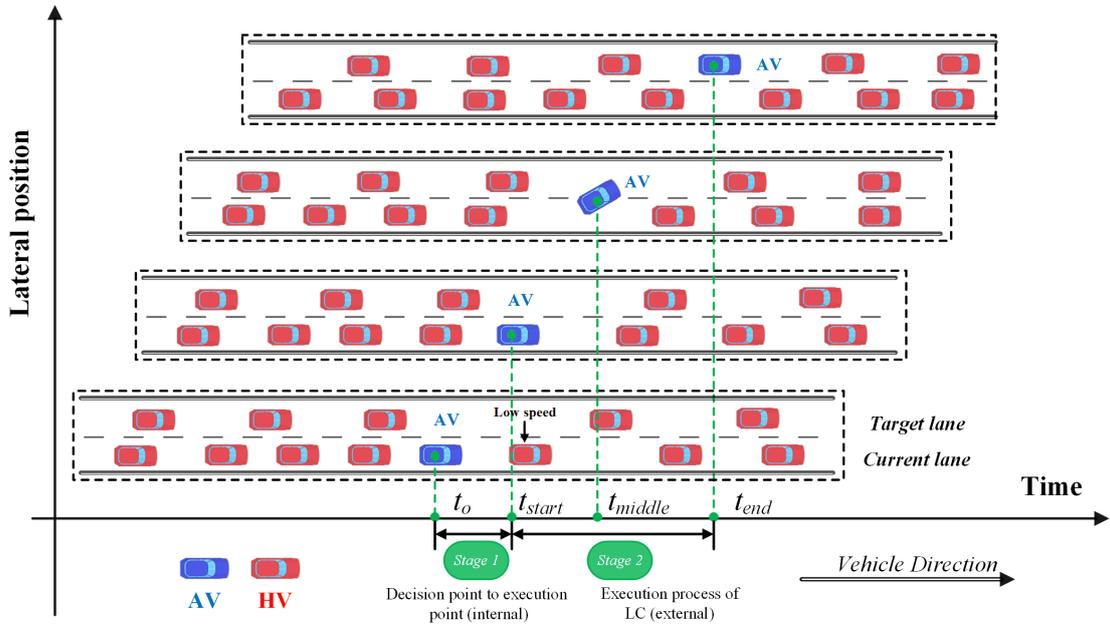

**Figure 2 Schematic diagram of the process of lane-changing maneuver**

The architecture of our proposed algorithm is given in Figure 3. In the upper layer, the joint optimization function of AV and the surrounding vehicles is formulated, which takes the comfort, efficiency, safety and fuel consumption into account. The quintic polynomial form is employed to model the LC trajectory of the AV, the LCM (Longitudinal Control Model) [27] is introduced to characterize the car-following behavior of HVs. The operational layer controller is composed of the vehicle kinematic model, vehicle error model, and the design of the MPC (Model Predictive Controller). Finally, these two modules are integrated together in the co-simulation platform, which is built with Matlab/Simulink and Carsim.

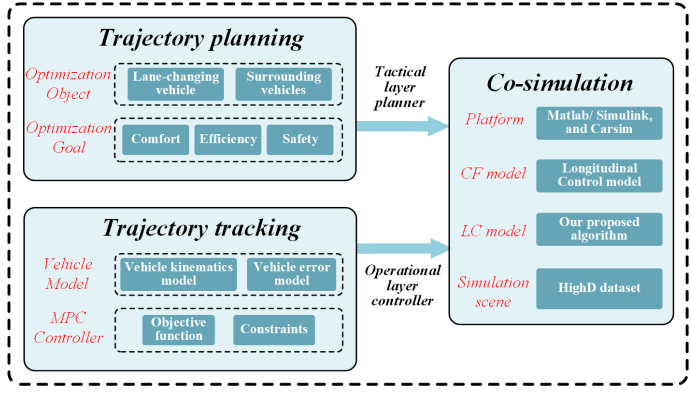

**Figure 3** Schematic diagram of the proposed hierarchical automatic LC algorithm

## 3 THE FORMULATION OF THE PROPOSED ALGORITHM
### 3.1 Cost function of AV

The time-based quintic polynomial function is introduced to model the longitudinal and lateral trajectory of the AV. This kind of function has been widely used in the existing research [21, 22], and exhibits better performance in fitting the LC trajectory. The longitudinal and lateral trajectory with respect to time is defined as below:

$$\begin{cases} x_{AV}(t) = a_0 + a_1 t + a_2 t^2 + a_3 t^3 + a_4 t^4 + a_5 t^5 \\ y_{AV}(t) = b_0 + b_1 t + b_2 t^2 + b_3 t^3 + b_4 t^4 + b_5 t^5 \\ \theta_{AV}(t) = ar\tan\left(\dot{y}_{AV}(t)/\dot{x}_{AV}(t)\right) \end{cases} \quad (1)$$

Where $x_{AV}(t), y_{AV}(t), \theta_{AV}(t)$ denote the longitudinal position, lateral position and course angle. $a_i, i=0,1,...5$ and $b_j, j=0,1,...5$ are the corresponding coefficients.

The brief notation of the LC trajectory is shown by the dotted line in Fig 4. Assuming time $t$, $x_{AV}(t)$, $\dot{x}_{AV}(t)$, $\ddot{x}_{AV}(t)$, $\Delta x_{AV}$, $\Delta \dot{x}_{AV}$ denotes the position, speed, acceleration, relative distance, and relative speed respectively. $T_{LCD} = t_{end} - t_{start}$ denotes the LC duration, and $D_o$ denotes the lane width. It is reasonable to assume that the velocity and acceleration of AV are desired to be zero at the start and end position in the lateral direction. Therefore, we could derive the following equations.

$$\begin{cases} x_{AV}(t_{start}) = 0, \dot{x}_{AV}(t_{start}) = v_{AV}^{start}, \ddot{x}_{AV}(t_{end}) = a_{AV}^{start} \\ x_{AV}(t_{end}) = x_{final}, \dot{x}_{AV}(t_{end}) = v_{AV}^{end}, \ddot{x}_{AV}(t_{end}) = a_{AV}^{end} \end{cases} \quad (2)$$

$$\begin{cases} y_{AV}(t_{start}) = 0, \dot{y}_{AV}(t_{start}) = 0, \ddot{y}_{AV}(t_{start}) = 0 \\ y_{AV}(t_{end}) = D_0, \dot{y}_{AV}(t_{end}) = 0, \ddot{y}_{AV}(t_{end}) = 0 \end{cases} \quad (3)$$

Where $v_{AV}^{start}$ denotes the initial speed, $v_{AV}^{end}$ denotes the final speed, $a_{AV}^{start}$ denotes the initial acceleration, $a_{AV}^{end}$ denotes the final acceleration, $x_{final}$ denotes the final longitudinal distance.

If we assume the AV has the same speed at the starting and ending point in the longitudinal direction. We could simplify the Equation (1) and obtain the following trajectory form.

$$\begin{cases} x_{AV}(t) = v_{AV}^{start}t - [v_{AV}^{start}T_{LCD} - x_{final}] \cdot z_{AV}(t) \\ y_{AV}(t) = 6D_o \cdot z_{AV}(t) \\ z_{AV}(t) = 6(t/T_{LCD})^5 - 15(t/T_{LCD})^4 + 10(t/T_{LCD})^3 \end{cases} \quad (4)$$

Where only two unknown parameters $T_{LCD}$ and $x_{final}$ are in the above formula.

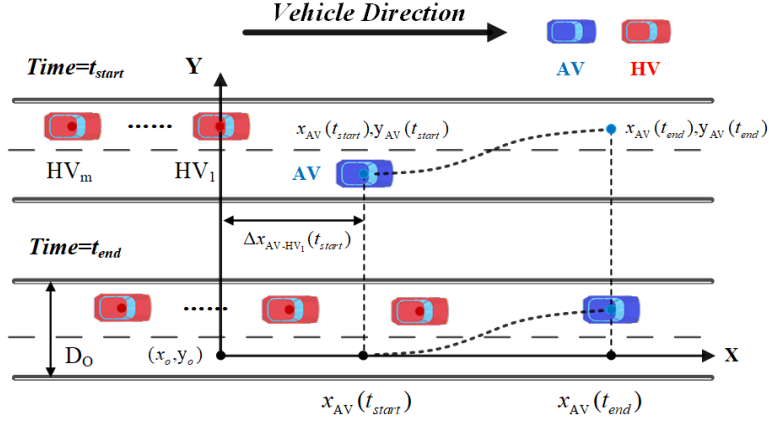

**Figure 4 The brief notation of the LTP algorithm**

The optimization objective function of AV consists of two parts: comfort and efficiency. The acceleration of AV should change smoothly rather than abruptly. Therefore, the variable, jerk, is introduced to serve as a measure of driver's comfort/discomfort.

$$L_{AV}^{comfort} = \sum_{t=t_{start}}^{t_{end}} |j_{AV,x}(t)| + \sum_{t=t_{start}}^{t_{end}} |j_{AV,y}(t)| \quad (5)$$

Where $j_{AV,x}(t)$, $j_{AV,y}(t)$ denote longitudinal and lateral jerk, maximum and minimum jerk/ acceleration.

The AV has to complete LC maneuver as soon as possible, so as to minimize the impact of LC on surroundings or avoid excessive occupation of surrounding road resources. Therefore, we introduce the difference between the current speed and desired speed.

$$L_{AV}^{efficiency} = \sum_{t=t_{start}}^{t_{end}} |v_{AV}(t) - v_{desire}| \quad (6)$$

Where $v_{AV}(t) = \sqrt{\dot{x}_{AV}(t)^2 + \dot{y}_{AV}(t)^2}$ and $v_{desire}$ denote the current and desired speed of AV.

Therefore, we could derive the total cost function of the AV at the given LC trajectory form.

$$Loss_{AV} = \omega_{AV}^{comfort} \cdot \frac{L_{AV}^{comfort}}{N_{AV}^{comfort}} + \omega_{AV}^{efficiency} \cdot \frac{L_{AV}^{efficiency}}{N_{AV}^{efficiency}} \quad (7)$$

Where $\omega_{AV}^{comfort}$, $\omega_{AV}^{efficiency}$ are the weight coefficient of the comfort and efficiency part. $N_{AV}^{comfort}$ and $N_{AV}^{efficiency}$ are the normalized values of the corresponding terms of AV in the objective function (to make the units consistent).

At the same time, the AV needs to meet the speed, stability, comfort, safety constrains. The speed of the AV should not exceed the maximum speed but should be greater than the minimum speed. The acceleration and jerk of AV should within the reasonable range.

$$\begin{cases} v_{min} \leq \sqrt{\dot{x}(t)^2 + \dot{y}(t)^2} \leq v_{max} \\ a_{min} \leq \ddot{x}(t) \leq a_{max} \\ a_{min} \leq \ddot{y}(t) \leq a_{max} \\ j_{min} \leq \dddot{x}(t) \leq j_{max} \\ j_{min} \leq \dddot{y}(t) \leq j_{max} \end{cases} \quad (8)$$

Where $v_{min}$, $v_{max}$ represent the minimum and maximum speed limit. $a_{min}$, $a_{max}$, $j_{min}$, $j_{max}$ denote the maximum acceleration, minimum acceleration, minimum jerk and maximum jerk respectively.

At the same time, the AV must not collide with surrounding vehicles at any time. The definition of the collision boundary area is shown below. The $l_a, l_b, C_a, C_b$ are defined as vehicle length, vehicle width, ellipse long radius and ellipse short radius respectively.

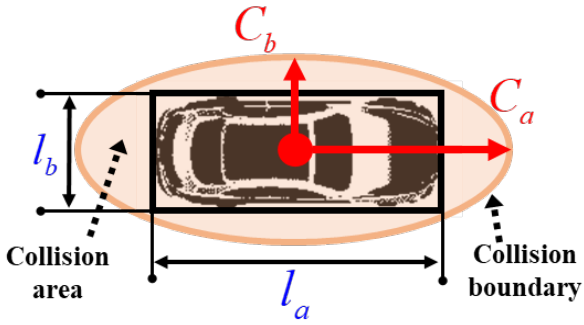

**Figure 5 The boundary of the collision area of the ego vehicle**

Taking the starting point as the original coordinates, suppose at time $t$, let $P_j(t) = (x_j(t), y_j(t))$ denotes the center position of the vehicle $j$. The real-time boundary of collision area of each vehicle is defined as $G_j(x, y)$.

$$\begin{cases} M^2/C_a^2 + N^2/C_b^2 = 1 \\ M = (x - x_j(t)) * \cos\theta_j - (y - y_j(t)) * \sin\theta_j \\ N = (x - x_j(t)) * \sin\theta_j + (y - y_j(t)) * \cos\theta_j \end{cases} \quad (9)$$

It is worth noting that the four corners of the smallest circumscribed rectangle of the vehicle outline should fall on the ellipse or within the ellipse. The real-time minimum distance between two collision boundaries could be obtained through the Lagrangian solution algorithm[17].

### 3.2 Cost function of HVs
Before giving the optimization objective function of HVs, we introduce the LCM model[27, 28] to characterize the longitudinal motions of HVs. The reason why we choose this model is to the unified perspective casted on the existing microscopic traffic flow models[28]. This model is derived through focusing the forces for the vehicle in the longitudinal direction of Field Theory. Field Theory represents everything in the environment (highways and vehicles) as a field perceived by the subject driver whose mission is to achieve his or her goals by navigating through the overall field. The formula is given below:

$$\ddot{x}_i(t+\tau_i) = A_i[1 - \frac{\dot{x}_i(t)}{v_{desire}} - e^{1-s_{ij}(t)/s_{ij}^*(t)}] \tag{10}$$

$$s_{ij}^*(t) = \frac{\dot{x}_i^2(t)}{2b_i} - \frac{\dot{x}_j^2(t)}{2B_j} + \dot{x}_i(t)\tau_i + L_j \tag{11}$$

Where $i$ denotes the follower vehicle, $j$ denotes the leader vehicle, $L_j$ denotes the vehicle length, $\tau_i$ denotes the reaction time, $s_{ij}(t)$ denotes the spacing between vehicle $i$ and $j$, $\Delta\dot{x}_{ij}(t)$ denotes the relative velocities, $v_{desire}$ denotes the desired speed, $A_i$ denotes the maximum acceleration, $s_{ij}^*$ denotes the desired safe spacing of driver $i$, $b_i$ denotes the maximum deceleration that the driver can confidently apply in an emergency, $B_j$ denotes the driver's estimation of the leader vehicle's comfortable deceleration in an emergency brake.

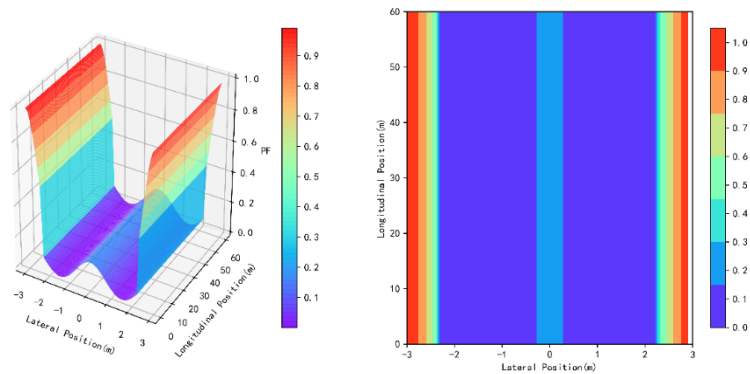

**Figure 6 Field of the road segment under the Field Theory**

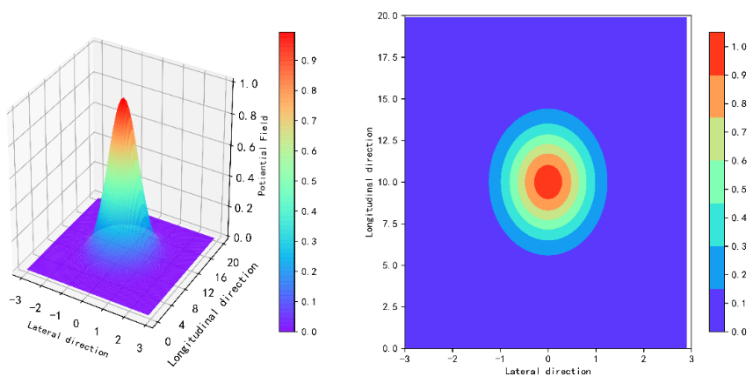

**Figure 7 Field of the subject vehicle under the Field Theory**

The optimization objective function of HVs contains the comfort and efficiency part. We also adopt the cumulative jerk as the comfort loss, the difference between the current and desired speed as the efficiency loss.

$$L_{HV_i}^{comfort} = \sum_{t=t_{start}}^{t_{end}} \left| j_{HV_i}(t) \right| \tag{12}$$

$$L_{HV_i}^{efficiency} = \sum_{t=t_{start}}^{t_{end}} \left| v_{HV_i}(t) - v_{desire} \right| \tag{13}$$

Where $j_{HV_i}(t)$ denotes the jerk of vehicle $HV_i$, which could be derived according to the Equation (13).

$$j_{HV_i}(t+\tau_i) = A_i \left[ -\frac{\ddot{x}_i(t)}{v_{desire}} - \frac{(s_{ij}^*(t))' s_{ij}(t)}{(s_{ij}^*(t))^2} e^{1-s_{ij}(t)/s_{ij}^*(t)} \right] \tag{14}$$

$$(s_{ij}^*(t))' = \frac{\ddot{x}_i(t)}{b_i} - \frac{\ddot{x}_j(t)}{B_j} + \ddot{x}_i(t)\tau_i \tag{15}$$

Where $(s_{ij}^*(t))'$ denote the derivative of the desired safe spacing with respect to time.

For vehicle with close distance and high relative speed difference, the more they are affected by the LC behavior, the larger its weight coefficient in the objective function. Therefore, we could derive the weight coefficient of each HV.

$$\omega_{HV_i} = \sigma_{HV_i} / \sum_{k=1}^{p} \sigma_{HV_k} \tag{16}$$

$$\sigma_{HV_i} = \left| \Delta \dot{x}_{HV_i}(t_{start}) \right| / \sqrt{\Delta x_{HV_i}(t_{start})} \tag{17}$$

Where $\Delta x_{H_iV}(t_{start})$ denotes the initial distance between $HV_i$ and AV, $\Delta \dot{x}_{HV_i}(t_{start})$ denotes the initial speed difference between $HV_i$ and AV.

Therefore, we could derive the total cost function of HVs as shown below.

$$Loss_{HV} = \omega_{HV}^{comfort} \cdot \frac{\sum_{i=1}^{p} \omega_{HV_i} L_{HV_i}^{comfort}}{N_{HV}^{comfort}} + \omega_{HV}^{efficiency} \cdot \frac{\sum_{i=1}^{p} \omega_{HV_i} L_{HV_i}^{efficiency}}{N_{HV}^{efficiency}} \tag{18}$$

Where $\omega_{HV}^{comfort}$ and $\omega_{HV}^{efficiency}$ denotes the corresponding weight coefficient of comfort, efficiency. $N_{HV}^{comfort}$, $N_{HV}^{efficiency}$ are the normalized values of the corresponding terms of HVs in the objective function (to make the units consistent).

Finally, we could derive the total loss of HVs and AV as shown below.

$$Loss_{total} = \omega_{AV} \cdot Loss_{AV} + \omega_{HV} \cdot Loss_{HV} \tag{19}$$

Where $\omega_{AV}$ represent the weight coefficient of the LC vehicle, which is between 0 and 1. $\omega_{HV} = 1 - \omega_{AV}$ represents the weight coefficient of HVs. It is worth noting that the consideration of comfort is based on safety, and the requirements for comfort are greater than safety. When $\omega_{AV}$ gradually approaches or equals to 1, AV would become pay more attention to its own loss, and cares less about the loss of surrounding vehicles. This optimization problem could be solved through various methods, like the SQP (Sequential Quadratic Programming) [22], IPA (Interior-Point Algorithm) [12], etc.

### 3.3 Vehicle kinematic modeling
Vehicle kinematics model considering the longitudinal, lateral, and yaw motions is introduced as shown in Fig 8. $(X_r, Y_r)$ and $(X_f, Y_f)$ are the coordinates of the center of the rear axle and the center of the front axle in the inertial coordinate system.

The state space of AV is defined as $\psi = [x, y, \varphi]$, and the input variable is defined $\mu = [v, \delta]^T$. According to the kinematic constraints of the front and rear axles, we could derive the following equations:

$$\dot{\psi} = [\dot{x}, \dot{y}, \dot{\varphi}]^T = [\cos\varphi, \sin\varphi, \frac{1}{l}\tan\delta]^T \cdot v_r \tag{20}$$

Where $\delta$, $l$, $v_r$ denotes the front wheel steering angle, wheel base, and the speed at the center of the rear axle. In order to facilitate the design of MPC (Model Predictive Controller), we expand this model by Taylor series at the reference trajectory point $(x_r, y_r)$, and we could derive the following vehicle error model.

$$\tilde{\psi}(k+1) = \tilde{A} \cdot \tilde{\psi}(k) + \tilde{B} \cdot \tilde{\mu}(k), (k=1,2,3\cdots) \tag{21}$$

$$\tilde{A} = \begin{bmatrix} 1 & 0 & -v_r\sin\varphi_r T \\ 0 & 1 & v_r\cos\varphi_r T \\ 0 & 0 & 1 \end{bmatrix}, \tilde{B} = \begin{bmatrix} \cos\varphi_r T & 0 \\ \sin\varphi_r T & 0 \\ \dfrac{\tan\delta_r T}{l} & \dfrac{v_r T}{l\cos^2\delta_r} \end{bmatrix} \tag{22}$$

Where $T$ denotes the sampling time, $\tilde{\psi} = \psi - \psi_r$ denotes the difference with the reference state, $\tilde{\mu} = \mu - \mu_r$ denotes the difference with the reference input, $\tilde{\psi}_r = [x_r, y_r, \varphi_r]^T$, $\mu_r = [a_r, \delta_r]^T$.

**3.4 Model predictive controller design**
The design ideas of the MPC controller mainly include: the current state should converge to the reference value as soon as possible, and the control input should as small as possible. Therefore, the deviation of the system state quantity and the control quantity need to be optimized. The objective function has the following form:

$$I(kq) = \sum_{i=1}^{N_p} \left\| \eta(k+i|t) - \eta_r(k+i|t) \right\|_R^2 + \sum_{i=1}^{N_c-1} \left\| \Delta U(k+i|t) \right\|_Q^2 + \rho\varepsilon^2 \tag{23}$$

Where the first item on the right side describes the rapidity of tracking control system, and the right side describes the stationarity of the tracking control system. $N_P$ is the prediction horizon, $N_C$ is the control horizon, and $\rho$ is the weight coefficient. $\varepsilon$ is the relaxation factor, which could directly limit the control increment, avoid the sudden change of control quantity, and prevent the situation that there is no feasible solution in the optimization process.

In the objective function, it is necessary to predict the output of the control system for a period of time in the future. The control output expression of the system is obtained by iterative derivation.

$$Y(t) = \lambda_t \psi(t) + \theta_t U(t) \tag{24}$$

Where $Y(t) = \begin{bmatrix} \tilde{\eta}(t+1) \\ \tilde{\eta}(t+2) \\ \cdots \\ \tilde{\eta}(t+N_p) \end{bmatrix}, \lambda_t = \begin{bmatrix} \tilde{A} \\ \tilde{A}^2 \\ \cdots \\ \tilde{A}^{N_p} \end{bmatrix}, \tilde{\psi}(t) = \begin{bmatrix} \tilde{X}_t \\ \tilde{Y}_t \\ \tilde{\varphi}_t \end{bmatrix},$

$\theta_t = \begin{bmatrix} \tilde{B} & 0 & 0 & \cdots & 0 \\ \tilde{A}\tilde{B} & \tilde{B} & 0 & \cdots & 0 \\ \tilde{A}^2 B & \tilde{A}\tilde{B} & \tilde{B} & \cdots & 0 \\ \vdots & \vdots & \vdots & \cdots & \vdots \\ \tilde{A}^{N_p-1}\tilde{B} & \tilde{A}^{N_p-2}\tilde{B} & \tilde{A}^{N_p-3}\tilde{B} & \cdots & 0 \end{bmatrix}, \Delta U(t) = \begin{bmatrix} \Delta \mu(t) \\ \Delta \mu(t+1) \\ \cdots \\ \Delta \mu(t+N_c) \end{bmatrix}$

## 4 SIMULATION EXPERIMENTS DESIGN AND RESULTS ANALYSIS
### 4.1 Metrics for evaluation and parameters settings

The total travel time is introduced to evaluate the improvement of the traffic operation. Total travel time (TTT) is the main indicator to assess the traffic operation and efficiency of transportation system. Through comparing the TTT difference for all vehicles to pass through the downstream road section under the two situations where the AV changes lane and the AV does not change lane, we could evaluate the impact of LC maneuver on the traffic flow. The cross section is selected 200 meters downstream from the starting point of LC.

For our proposed hierarchical trajectory planning and tracking algorithm, their common parameters use the same values respectively. Several parameters related to the trajectory planning mainly refer to our previous research [17]. The simulation step is about 0.1s. $l_a = 5m$, $l_b = 2m$, $C_a = 2.5m$, $C_b = 1m$, $v_{max} = 30m/s$, $v_{min} = 5m/s$, $a_{max} = 8m/s^2$, $a_{min} = -8m/s^2$, $j_{max} = 8m/s^3$, $j_{min} = -8m/s^3$, $D_o = 3.5m$ [17]. The parameters of the LCM model are: $A_i = 2.81m/s^2$, $b_i = 6.14m/s^2$, $B_j = 5.95m/s^2$, $\tau_i = 0.46s$, $v_{desire} = 25m/s$, $L_j = 5.03m$ [29]. $N_P = 6$, $N_c = 4$. $N_{AV}^{comfort} = N_{HV}^{comfort} = j_{max}$, $N_{AV}^{efficiency} = N_{HV}^{efficiency} = v_{desire}$.

### 4.2 Testing scenario 1: comparison with benchmark algorithm

To show the advantage of our proposed algorithm, we compare this with the benchmark algorithm. The benchmark algorithm is consistent with the algorithm in the existing literature, which not considers the loss function of surrounding vehicles in the optimization objective function ($\omega_{AV} = 1$).

● **Benchmark scenario settings:** the total simulation time is 300s, and AV performs LC at 80s, with a distance of 10 meters from the leader vehicle $HV_1$. The comfort and efficiency weight coefficients are all 0.5, and the weight coefficient of AV is 0.5. The speed of $HV_1$, and the initial and target speed of AV are all $25m/s$. For simplicity, we assume 10 vehicles are behind the target lane (we will set up a more complex scene later). Figure 8 presents the trajectory information of HVs. Four subgraphs represent the longitudinal position, speed, acceleration and time-headway of HVs.

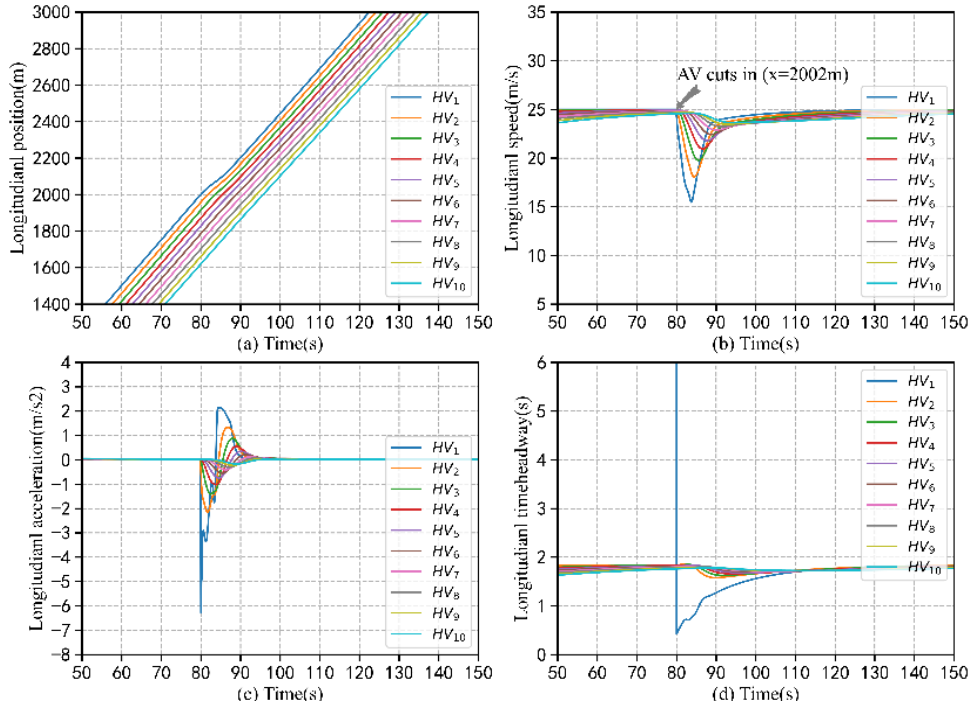

**Figure 8 Trajectory information of HVs under the benchmark scenario**

● **Results of the benchmark algorithm:** due to the LC behavior of AV, the speed of following HVs exhibits an obvious deceleration. The speed of $HV_1$ suddenly reduced from 25m/s to 15.47m/s, and the time-headway of $HV_1$ reduced to 0.42s. The loss of AV has reached its own lowest value, the corresponding comfort and efficiency loss is about 28.87 and 13.97. At the same time, the comfort and efficiency loss of HVs is about 27.50 and 29.79. The total loss of AV and HVs is about 50.06 as shown in Figure 10. Under this scenario, the total TTT difference is about 14.18s.

● **Results of our proposed local-optimum algorithm:** (the loss of HVs is considered in the optimization objective function of AV): the total loss of HVs and AV decreased from 50.06 to 44.64. The total loss of HVs decreases from 28.64 to 17.70. The AV has made a certain degree of compromise, whose total loss increases from 21.42 to 26.94. The increase in loss of AV is lower than the decrease in loss of HVs. Under this scenario, the total TTT difference has also been decreased from 14.18s to 11.84s. This indicates that the overall operating status of the traffic flow within the LC area has been improved.

● **Comparison of the LC trajectory:** Figure 9 presents the comparison of the LC trajectory under the benchmark and our proposed local-optimum algorithm, which shows the position, speed, and acceleration both in the longitudinal and lateral direction. Under our algorithm, the loss of the following HVs are considered in the optimization objective function. Therefore, AV chooses the strategy of accelerating in the longitudinal direction. Compared with the benchmark scenario, the speed of AV gradually increases from 25m/s to the highest 28.11m/s, and then decrease to 25m/s. And the AV takes a smaller acceleration in the lateral direction. The maximum lateral speed difference is about 0.2m/s.

● **Sensitivity analysis of** $\omega_{AV}$**:** Figure 11 presents the total loss of AV and HVs under different value of $\omega_{AV}$, where we vary the value of $\omega_{AV}$ from 0.1 to 1. High value of $\omega_{AV}$ indicates that AV pays more attention to its self-interest, while low value of $\omega_{AV}$ indicates that it cares more about its impact on surrounding vehicles. With the gradual increase of $\omega_{AV}$, the cost of AV gradually decreases from 36.79 to 21.42, and the cost of HVs increases from 14.14 to 28.64. The total cost exhibits a trend of first decreasing and then increasing, reaching the lowest value when $\omega_{AV}$ equals to 0.5 or 0.6. Results demonstrates that we could indeed obtain a LC trajectory that achieves a local optimum, which could minimize the total loss within the LC area, thus improving the overall comfort and efficiency.

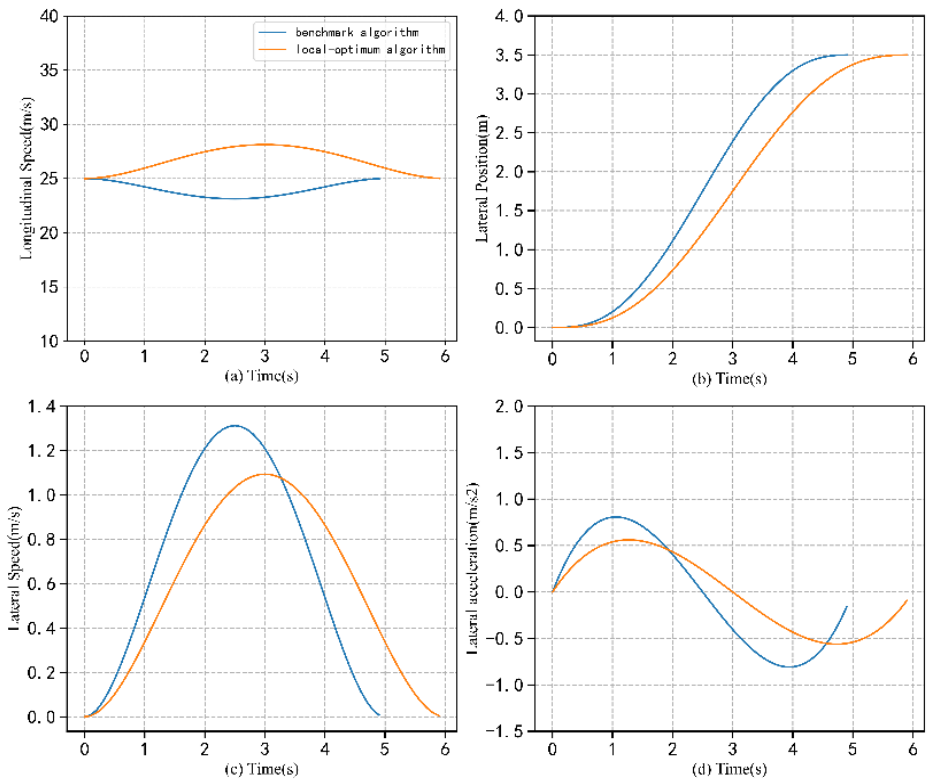

**Figure 9 Comparison of the LC trajectory between the benchmark scenario and optimal scenario**

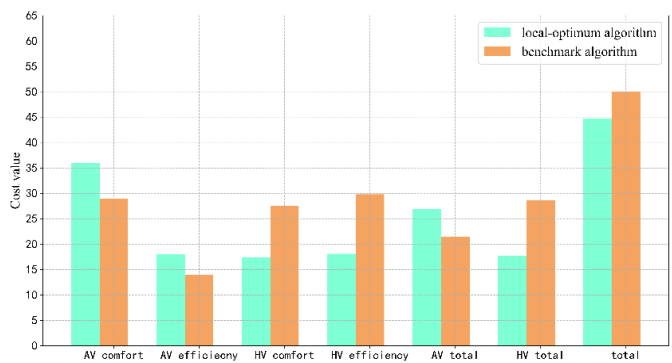

**Figure 10 Comparison of the loss between the benchmark scenario and optimal scenario**

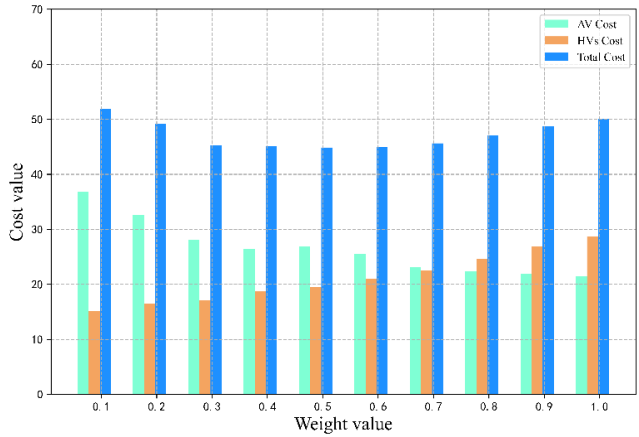

**Figure 11 Sensitivity analysis of the coefficient $\omega_{AV}$ in the objective function**

**4.3 Testing scenario 2: comparison of the traffic shock wave**
This subsection presents the comparison of the traffic shock wave caused by the LC maneuver. The total simulation time is about 800s, and the LC time is 200s. A number of 100 HVs are set, and the speed of the initial preceding is 25m/s. The initial and target speed of AV is set as 20m/s and 15m/s respectively (to verify the performance of our algorithm under extreme conditions).

● **Traffic shock wave impact analysis:** Figure 12and Figure 13 presents the speed time-space heatmap of HVs under different cut-in speed of AV. Through the heatmap, we could not only see the formation of traffic shock wave, but also the corresponding speed changes. The abscissa represents the simulation time, the ordinate represents the longitudinal position, and each curve represents the trajectory of the vehicle. It can be seen that when AV is inserted in front of the HVs at a low speed, a corresponding traffic shock wave is formed. The closer the color is to blue, the higher the speed of HVs. The closer the color is to red, the lower the speed of HVs. The blue area represents the speed around 20~25m/s, the red area represents the speed around 15~20 m/s. Three different subgraphs are exhibited. The subgraphs from left to right are our proposed local-optimum algorithm, benchmark algorithm, and the speed difference between these two algorithms. Since the LC does not last for a longer duration, the traffic shock wave mainly has obvious differences in the green range (We will study this area in depth later). It can be obviously found that the speed of the vehicle in the green box on the left subgraph is higher than that in the middle subgraph. This is reflected in the red range is smaller. This indicates that our proposed algorithm could indeed enhance the speed of the vehicles within the LC area, even though the scope of the impact of LC is not particularly large. The speed of the immediately-following vehicles increases by about 1~5m/s as shown in the right subgraph, and the speed of rear 5 to 10 vehicles has increased most significantly.

● **Flow-density-speed comparison:** Since the speed of AV during the LC is not a fixed value, it is not appropriate for us to estimate the traffic shock wave using the steady-state macroscopic equivalent of LCM [27]. Nevertheless, we could employ the basic formulas to explore the flow, space-mean speed, and traffic density within the LC area [30]. In order to further analyze the impact of the above traffic shock wave, we extract the HV trajectories within the red box as shown in Figure 14. We employ the method in [30] to estimate the flow, space-mean speed, and density within this area, according to the following three formulas.

$$q_{flow} = d(A)/|A| \qquad (25)$$
$$V_{space-mean} = d(A)/t(A) \qquad (26)$$
$$k_{density} = t(A)/|A| \qquad (27)$$

Where $|A|$ denotes the area of rectangle, $d(A)$ and $t(A)$ denotes the total distance and time travelled of all vehicles within this area.

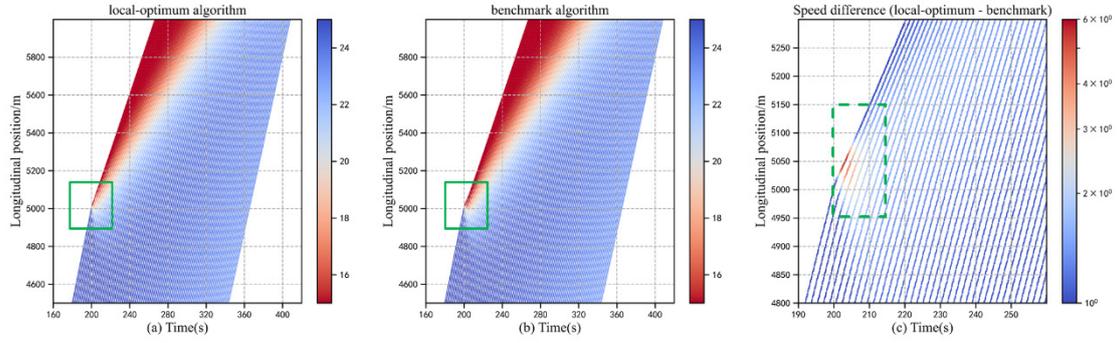

**figure 12 Traffic shock wave caused by the LC maneuver (initial and target speed of AV is 15m/s)**

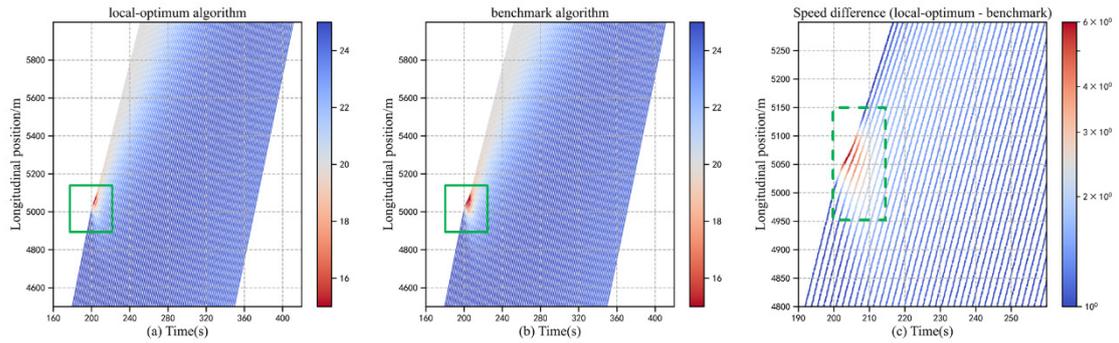

**Figure 13 Traffic shock wave caused by the LC maneuver (initial and target speed of AV is 20m/s)**

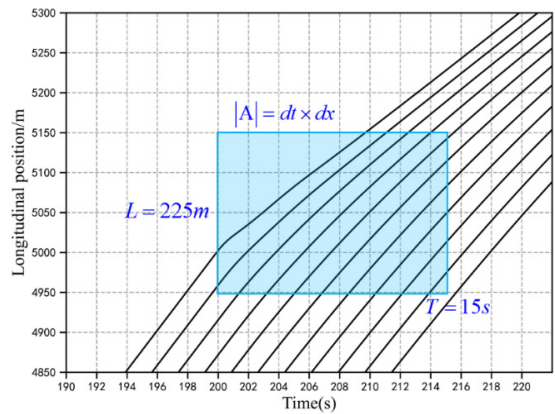

**Figure 14 Vehicle trajectories within the area of rectangle**

**Table 1 Time and distance travelled of each vehicle within the LC area under initial different speed of AV**

| Vehicle | initial and target speed of AV is 15m/s | | | | initial and target speed of AV is 20m/s | | | |
|---|---|---|---|---|---|---|---|---|
| | Time travelled(s) | | Distance travelled(m) | | Time travelled(s) | | Distance travelled(m) | |
| | bench mark | local-optimum | bench mark | local-optimum | bench mark | local-optimum | bench mark | local-optimum |
| $HV_1$ | 10.80 | 9.80 | 147.22 | 147.48 | 9.80 | 8.70 | 148.35 | 148.44 |
| $HV_2$ | 12.10 | 11.20 | 191.36 | 191.43 | 11.00 | 10.10 | 191.45 | 191.75 |
| $HV_3$ | 13.40 | 12.60 | 234.41 | 234.42 | 10.80 | 11.60 | 197.33 | 235.56 |
| $HV_4$ | 13.60 | 12.90 | 248.25 | 248.43 | 10.50 | 9.80 | 199.29 | 198.48 |
| $HV_5$ | 12.10 | 12.10 | 229.08 | 239.08 | 10.00 | 9.60 | 196.47 | 199.57 |
| $HV_6$ | 10.30 | 10.40 | 202.78 | 213.10 | 8.40 | 8.20 | 170.39 | 174.59 |
| $HV_7$ | 9.00 | 8.60 | 184.97 | 181.65 | 6.30 | 6.40 | 130.99 | 139.05 |
| $HV_8$ | 6.80 | 6.80 | 143.16 | 147.24 | 4.60 | 4.70 | 97.77 | 103.69 |
| $HV_9$ | 5.00 | 5.00 | 107.71 | 110.26 | 2.80 | 2.90 | 60.08 | 64.24 |
| $HV_{10}$ | 3.20 | 3.30 | 69.78 | 73.48 | 1.10 | 1.20 | 22.72 | 25.63 |
| Total | 96.30 | 92.70 | 1758.72 | 1786.56 | 75.30 | 73.20 | 1414.82 | 1481.01 |

**Table 2 Flow, space-mean speed, and traffic density under different initial speed of AV**

| | initial and target speed of AV is 15m/s | | initial and target speed of AV is 20m/s | |
|---|---|---|---|---|
| | benchmark | local-optimum | benchmark | local-optimum |
| Traffic flow (veh/h) | 1875.963 | 1905.666 | 1509.140 | 1579.746 |
| Space-mean speed (km/h) | 65.746 | 69.381 | 67.641 | 72.837 |
| density (veh/km) | 28.533 | 27.467 | 22.311 | 21.689 |
| Total loss of HVs and AV | 224.96 | 207.58 | 198.16 | 184.65 |

Table 1 presents the time and distance travelled of each vehicle within the LC area under different initial speed of AV. $|A|$ equals to $L \times T = 225 \times 15 = 3375 m \cdot s$. According to the above formula, we could derive the flow, space-mean speed, and density as shown in Table 2. It could be found that when the initial speed of is 15m/s, the traffic flow between the optimal and benchmark scenario is about 1905veh/h and 1875.96veh/h. the space-mean speed is about 65.75km/h and 69.38km/h. Under our proposed algorithm, the traffic flow increases by 30.30veh/h, and speed increases by 3.64m/s. At the same time, from a micro point of view, we found that the total loss within the LC area shows the tendency of decreasing. The total loss reduces from 224.96 to 207.58. When the initial speed of AV becomes 20m/s, the traffic flow increases from 1509.14veh/h to 1579.75veh/h, the speed increases from 67.64m/s to 72.84m/s, and the total loss reduces from 198.16 to 184.65. These results all indicate that our algorithm could obviously enhance the traffic efficiency within the LC area.

## 4.4 Testing scenario 3: demonstration of the application of the algorithm

In this section, we present the demonstration of application of this algorithm to real-traffic environment. Although there is no vehicle trajectory data for fully autonomous driving, it is at least appropriate for us to use the naturalistic driving data to examine the effectiveness of the algorithm. We attempt to extract field-data LC and CF trajectories from the HighD dataset [21], and utilize the initial trajectory information as the input of our proposed algorithm. The details of the extraction process of LC could be found in our recent research [6]. After obtaining the complete LC trajectories, we could then obtain the CF trajectories according to the ID information of the target following vehicle. Figure 15 presents the example of LC trajectory. ID27, ID26, ID30, ID23 denote the LC vehicle, current preceding, current following, and target preceding vehicle. ID28, ID31, ID 35 denotes three following vehicles on the target lane.

● **Calibration of the LCM model:** we employ the CF trajectory data before LC to calibrate the LCM model. The calibration process of CF model is designed to minimize the difference between the actual CF trajectory and the simulated trajectory data generated by the CF model. For each trajectory, we input the actual velocities of the follower and the leader, and the actual spacing distance into the CF model. We set the initial population to 200, the maximum iteration to 200, the intersection probability to 0.95 and the mutation probability to 0.05 [29]. The final objective function value is about 0.09, and the parameters after calibration are given in Table 3.

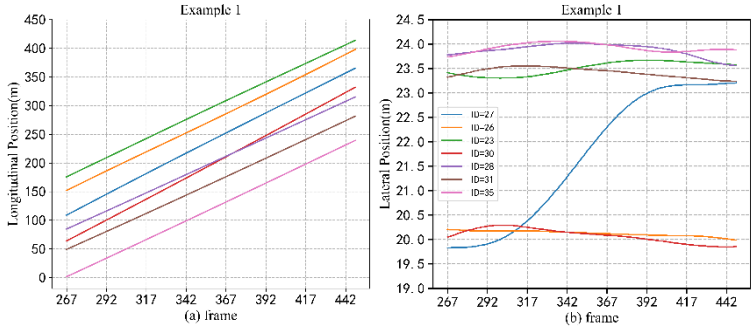

**Figure 15 Example of the LC and CF trajectory (three rear vehicles are on the target lane)**

**Table 3 Calibration results of the LCM model**

| Notation | Range [min, max] | 25th quantile | 50th quantile | 75th quantile | mean |
|---|---|---|---|---|---|
| $v_{desire}$ | [5,50] | 31.81 | 33.00 | 34.91 | 33.54 |
| $A_i$ | [2,10] | 4.18 | 4.44 | 5.21 | 4.91 |
| $\tau_i$ | [0.1,10] | 0.56 | 0.82 | 1.31 | 0.93 |
| $L_j$ | [2,10] | 4.64 | 4.80 | 4.91 | 4.82 |
| $b_i$ | [2,10] | 6.26 | 6.46 | 8.19 | 6.93 |
| $B_j$ | [2,10] | 5.75 | 6.34 | 7.30 | 6.44 |

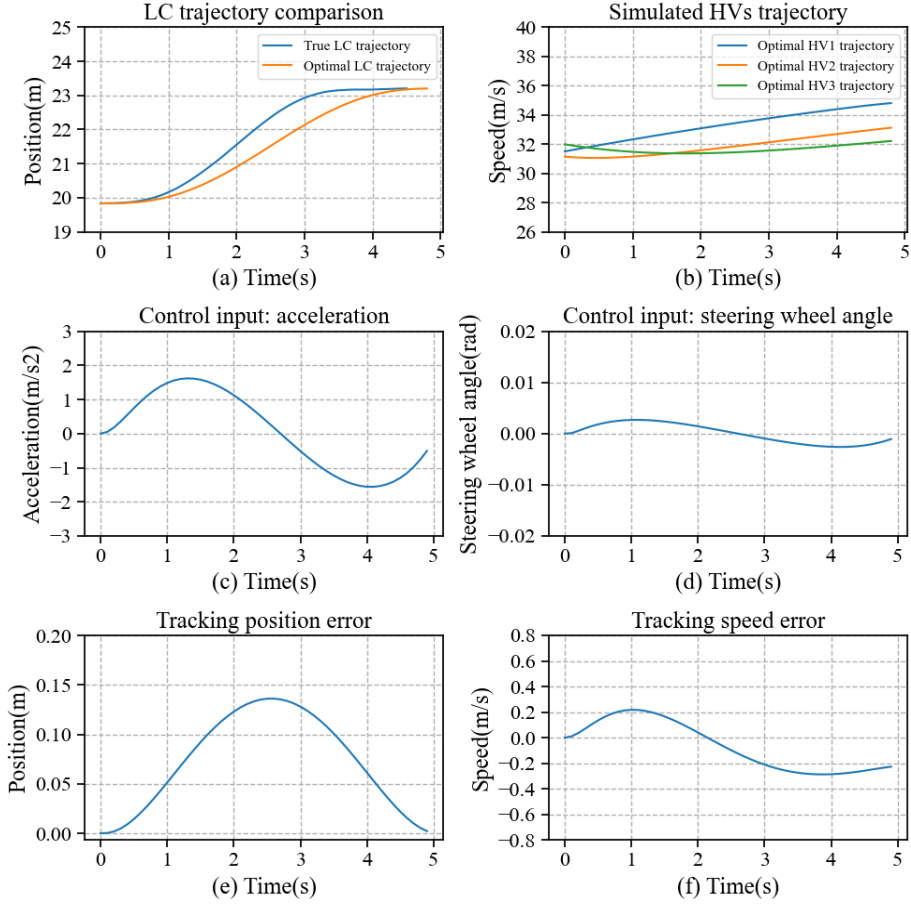

**Figure 16 Comparison of the simulation results (LC trajectory and control input)**

● **Parameter settings:** the desired speed is set as 33.54m/s, the weight coefficient of AV is 0.5, and the simulation step is set as 0.1s. For the LCM model, we employ the parameters in Tab. 4. The initial distance and speed difference between the LC vehicle and three following vehicles on the target lane are 24.23m, 59.73m, 107.26m, and 5.27m/s, 5.43m/s, 4.56m/s respectively. According to Eq. 17, the weight coefficients of three HV vehicle are 0.48, 0.32 and 0.2. The initial distance and speed difference with current preceding vehicle are about 43.13m and 2.32m/s. The initial gap distance on the target lane is about 91.07m. We assume that current preceding, current following, and target preceding vehicle are also under the control of LCM.

● **Results analysis:** Figure 16 presents the simulation results. Subgraph (a) presents the comparison between the true and the simulated LC trajectory. Subgraph (b) presents the simulated HVs trajectories. After obtaining the results of upper layer, AV is controlled under the MPC controller. Subgraph (c) to subgraph (d) present the control input of the AV, and the tracking error of speed and position. Meanwhile, under the benchmark scenario, the loss of AV is about 21.82, the loss of HVs is about 48.44, and the total loss is about 70.28. While under our algorithm, the loss of AV increase from 21.84 to 25.4, the loss of HVs decreases from 48.44 to 42.12. The total loss decreases from 70.28 to 67.52. Results also demonstrate that under our proposed algorithm could not only ensure the safe completion of LC maneuver, but also minimize the total loss within the LC area.

## 4.5 Discussion

As mentioned in the literature review, the center of this paper is not only the proposition of the novel algorithm, but also the application of the concept of local-optimum LC algorithm. We aim to actively dissipate or eliminate traffic shock waves caused by the LC maneuver from the source as much as possible. Although existing research has achieved fruitful results, they often ignore the loss of the immediately-following vehicles behind the target lane. Consequently, the planned trajectory might indeed optimal for the LC vehicle, while the overall loss in the LC area might not minimal. Therefore, this paper has proposed a novel automatic LC algorithm that could achieve local-optimum within the LC area. Not only the safe completion of LC maneuver is ensured, but also the impact of LC maneuver on surrounding vehicles is minimized. In the tactical layer, the loss of the LC vehicle and the following vehicles are considered in the optimization objective function of the LC trajectory of AV. The comfort, safety, and efficiency of the LC vehicle and surrounding vehicles are simultaneously satisfied in the optimization objective function. In the operational layer, the vehicle kinematic model and the design of MPC controller is employed to track the generated LC trajectory.

Combining macro-level and micro-level analysis, we explore the effectiveness of our proposed algorithm. From the micro level, we analyze the loss of each vehicle, the total loss within the LC area, and the trajectory for each vehicle in detail. Since the LC maneuver only lasts for a few seconds, we narrow the research area, and study the traffic flow state within the LC area. From the macro level, we explore the traffic shock wave caused by LC maneuver. We roughly estimate the flow, space-mean speed, density within the LC area, through employing the methods in [30]. Results demonstrate that compared with the benchmark algorithm, the impact of LC behavior of AV on traffic flow can be significantly reduced. This is reflected in the decrease of the total loss of vehicles within the LC area, and the increase of the space-mean speed around the LC area. These findings suggest that our proposed algorithm has indeed improved the state of traffic flow within the area, even though this area is rather small ($|A| = 3375 m \cdot s$). To guide the practical application of our algorithm, we employ the trajectory information from the HighD dataset to validate the effectiveness of the algorithm.

Undoubtedly, many aspects of this paper need further research. As elaborated in the problem statement, we mainly focus on obtaining a local-optimum LC trajectory for the AV. This is an important issue for future research. The determination of the execution point is also an indispensable component of automatic LC algorithm. The determination of the local-optimal execution point, and the combination of these two stages together may further reduce the impact of LC behavior on traffic flow. Our current research could be regarded as the most preliminary work of these two parts, and it is difficult for us to bypass this research if we want to conduct these two parts. The ongoing research is to combine the traffic wave theory to construct the optimal decision point. Through integrating the steady-state macroscopic CF model [27] and the LWR [31, 32] model, we may could derive the optimal execution point with the smallest shock wave speed. On the other hand, considering that there will may be scenes where AVs and HVs are mixed on the road, establishment of a locally-optimal automatic LC algorithm for this situation also needs to be researched.

## 5 CONCLUSION

This paper proposes a hierarchical LC motion planning algorithm, which could make the traffic flow reach a local-optimal state. The loss function of AV and HVs is formulated in the tactical layer, and the operational layer consists of the vehicle kinematic model and MPC controller. We evaluate the proposed algorithm both from the macro and micro perspective. To guide the practical application of our algorithm, we extract LC and CF trajectories from the HighD dataset, and utilize the initial data as the input. Results all demonstrate that our proposed algorithm could improve the overall performance of traffic flow within the LC area. We hope this research could promote the study of automatic LC algorithm, especially for the local-optimum LC algorithm.

## 6 AUTHOR CONTRIBUTION STATEMENT

The authors confirm contribution to the paper as follows: Yang Li: Conceptualization, Data curation, Writing - original draft. Linbo Li: Methodology, Funding acquisition, Writing - original draft. Daiheng Ni: Investigation, Writing - review & editing, but with no involvement in the research grant. Wenxuan Wang: Conceptualization and Writing.

## 7 ACKNOWLEDGEMENTS

This research was funded by the National Key R&D Program of China [grant numbers 2018YFE0102800].